\documentclass[10pt,twocolumn,letterpaper]{article}

\usepackage[pagenumbers]{cvpr} 

\usepackage{graphicx}
\usepackage{amsmath}
\usepackage{amssymb}
\usepackage{booktabs}
\usepackage{textcomp}
\usepackage{tabularx}
\usepackage{threeparttable}

\usepackage[pagebackref,breaklinks,colorlinks]{hyperref}
\usepackage{enumitem}
\newcommand\nnfootnote[1]{%
  \begin{NoHyper}
  \renewcommand\thefootnote{}\footnote{#1}%
  \addtocounter{footnote}{-1}%
  \end{NoHyper}
}

\usepackage[capitalize]{cleveref}
\crefname{section}{Sec.}{Secs.}
\Crefname{section}{Section}{Sections}
\Crefname{table}{Table}{Tables}
\crefname{table}{Tab.}{Tabs.}

\def\cvprPaperID{663} 
\def\confName{CVPR}
\def\confYear{2022}

\begin{document}

\title{PointMatch: A Consistency Training Framework for  Weakly Supervised Semantic Segmentation of 3D Point Clouds}

\author{Yushuang Wu$^{123*}$\quad Zizheng Yan$^{123*}$\quad Shengcai Cai$^{13}$\\Guanbin Li$^{43}$\quad Yizhou Yu$^{56}$\quad Xiaoguang Han$^{123\dag}$\quad Shuguang Cui$^{123}$\\
$^1$SSE, CUHKSZ\quad $^2$FNii, CUHKSZ\quad $^3$SRIBD\quad $^4$SYSU\quad $^5$HKU\quad $^6$DeepWise\\
{\tt\small \{yushuangwu, zizhengyan, shengcaicai\}@link.cuhk.edu.cn} \\ {\tt\small liguanbin@mail.sysu.edu.cn\quad yizhouy@acm.org\quad  \{hanxiaoguang, shuguangcui\}@cuhk.edu.cn}}

\maketitle

\begin{abstract}
   Semantic segmentation of point cloud usually relies on dense annotation that is exhausting and costly, so it attracts wide attention to investigate solutions for the weakly supervised scheme with only sparse points annotated. Existing works start from the given labels and propagate them to highly-related but unlabeled points, with the guidance of data, e.g. intra-point relation. However, it suffers from (i) the inefficient exploitation of data information, and (ii) the strong reliance on labels thus is easily suppressed when given much fewer annotations. Therefore, we propose a novel framework, PointMatch, that stands on both data and label, by applying consistency regularization to sufficiently probe information from data itself and leveraging weak labels as assistance at the same time. By doing so, meaningful information can be learned from both data and label for better representation learning, which also enables the model more robust to the extent of label sparsity. Simple yet effective, the proposed PointMatch achieves the state-of-the-art performance under various weakly-supervised schemes on both ScanNet-v2 and S3DIS datasets, especially on the settings with extremely sparse labels, e.g. surpassing SQN by 21.2\% and 17.2\% on the 0.01\% and 0.1\% setting of ScanNet-v2, respectively. 
   \nnfootnote{$^*$Equal contribution}
   \nnfootnote{$^\dag$Corresponding author}
\end{abstract}

\section{Introduction}
Semantic segmentation of 3D point clouds is crucial for the application of intelligent robots' understanding scenes in the real world. Great efforts have been contributed to the fully supervised scheme, while it requires exhausting and costly per-point annotations (\eg around 22.3 minutes to annotate an indoor scene on average~\cite{dai2017scannet}). Thus, weakly supervised 3D semantic segmentation now receives increasing attention, where only limited point-level annotations are provided in each point cloud.  

\begin{figure}[t]
\centering
\includegraphics[width=1.0\linewidth]{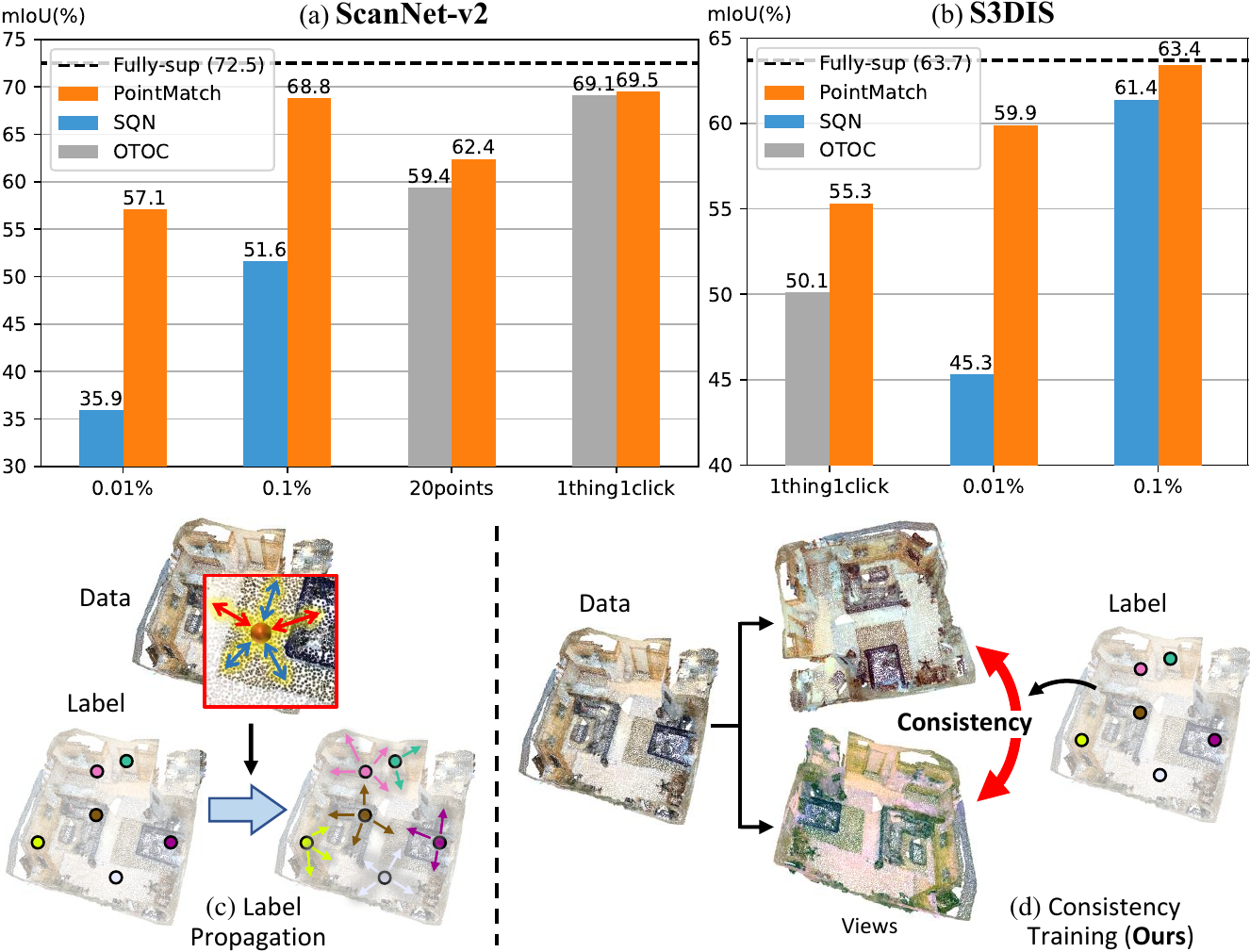}
\caption{(a), (b) the performance of PointMatch on the ScanNet-v2 and S3DIS datasets over various weakly supervised semantic segmentation settings: annotating 0.01\%, 0.1\% of points~\cite{hu2021sqn}, 20 points per-scene~\cite{hou2021exploring}, and ``1thing1click''~\cite{liu2021one}. (c), (d) a comparison between previous works and the proposed approach. }
\label{fig:teaser}
\vspace{-4mm}
\end{figure}

Recently, several approaches are proposed for weakly supervised point cloud semantic segmentation with different kinds of weak labels, including projected 2D image~\cite{wang2019towards}, subcloud-level~\cite{wei2020multi}, segment-level~\cite{tao2020seggroup}, and point-level~\cite{xu2020weakly, hou2021exploring, liu2021one, hu2021sqn} supervision. In this paper, we focus on addressing the setting of sparse point-level labels, which is one of the most convenient annotation schemes in the application. 
The key challenge of this task is the difficulty of learning a robust model given very sparse supervision in the point cloud (\eg 0.1\%, 0.01\% of points annotated in \cite{hu2021sqn} and around 0.02\% in \cite{liu2021one}). Existing solutions are mainly committed to alleviating the label sparsity by reusing limited supervision, \ie, first probing the highly-related points~\cite{hu2021sqn} or super-voxels~\cite{liu2021one} and allowing them to share the same training labels. However, this line of works are explicitly constructed on label propagation and employ point cloud data as the propagation guidance, which suffers from (i) the insufficient exploitation of data information limits the learning efficiency, and (ii) the propagated labels strongly rely on the original annotation scale thus the performance is easily suppressed when given much fewer labels. Therefore, we propose to probe information from both label and data itself for more efficient and robust representation learning. 

Recently, consistency training is acknowledged as a powerful algorithmic paradigm for robust learning from label-scarce data, \eg in unsupervised/semi-supervised learning~\cite{hu2017learning, grill2020bootstrap, xie2020self,  sohn2020fixmatch} and unsupervised/semi-supervised domain adaptation~\cite{french2017self, shu2018dirt, li2021cross, li2021semi}. It works by forcing the model to make consistent prediction under different perturbations/augmentations to the input sample (named as different \textit{views}) and the prediction in one view usually serve as the pseudo-label of the other view. Inspired by this, we propose a novel consistency training framework, PointMatch, for the weakly supervised 3D semantic segmentation. Given a whole scene of point cloud with sparse labels, PointMatch employs the per-point prediction in one view as the other's pseudo-label to encourage the predictive consistency between two views of a scene. Such consistency facilitates (i) robustness to easily-perturbed low-level input feature and (ii) stronger capability in learning useful high-level representations to keep predictive consistency. Besides, the provided labels act as extra supervision to assist high-level semantic feature discrimination, which also benefits the representation learning from data. By doing so, the reliance on the given label is relieved and more information is probed from the point cloud data itself. 

Originating from the per-point prediction in one view, the pseudo-label should be of high quality to provide positive guidance for the other. Whereas there exist considerable mispredictions especially at the early learning stage. Thus, we exploit the inherent structure of the point cloud to improve the pseudo-label quality, via integrating the super-point grouping information where similar points are clustered by low-level features (\eg position and color) into the same group and are assumed to have the same semantic meaning. Specifically, the grouping information is used to correct the minor predictions that diverge from the ``mainstream'' in the super-point. Despite its good property, the super-point-aware pseudo-label actually introduces noise from the pretext super-point generation. Therefore, to fully utilize these two types of pseudo-labels, we design an adaptive pseudo-labeling mechanism, where the model is encouraged to believe the super-point-aware pseudo-label more at the beginning, and gradually resorts to its raw prediction when the model itself is reliable enough. 
Extensive experiments and analysis on the ScanNet-v2~\cite{dai2017scannet} and S3DIS~\cite{armeni20163d} dataset validate the effectiveness of the proposed approach. As shown in Fig.~\ref{fig:teaser}, the proposed PointMatch significantly surpasses the state of the art on various weakly supervised schemes and impressively, shows great robustness given extremely sparse labels.   

The contributions of this paper are listed as follows:
\begin{itemize}[leftmargin=*]
\vspace{-2mm}
\item We propose a novel consistency training framework, PointMatch, for the weakly supervised 3D semantic segmentation, which can facilitate the network to learn robust representation from sparse labels and point cloud data. 
\vspace{-2mm}
\item We introduce super-point information to promote the pseudo-label quality in our framework, and it is employed in an adaptive manner to well utilize the advantages of both two types of pseudo-label.
\vspace{-2mm}
\item Extensive experiments validate the effectiveness and superiority of PointMatch, and the proposed approach achieves significant improvements beyond the state of the art over various weakly-supervised settings.
\end{itemize}

\begin{figure*}[t]
\centering
\includegraphics[width=0.85\linewidth]{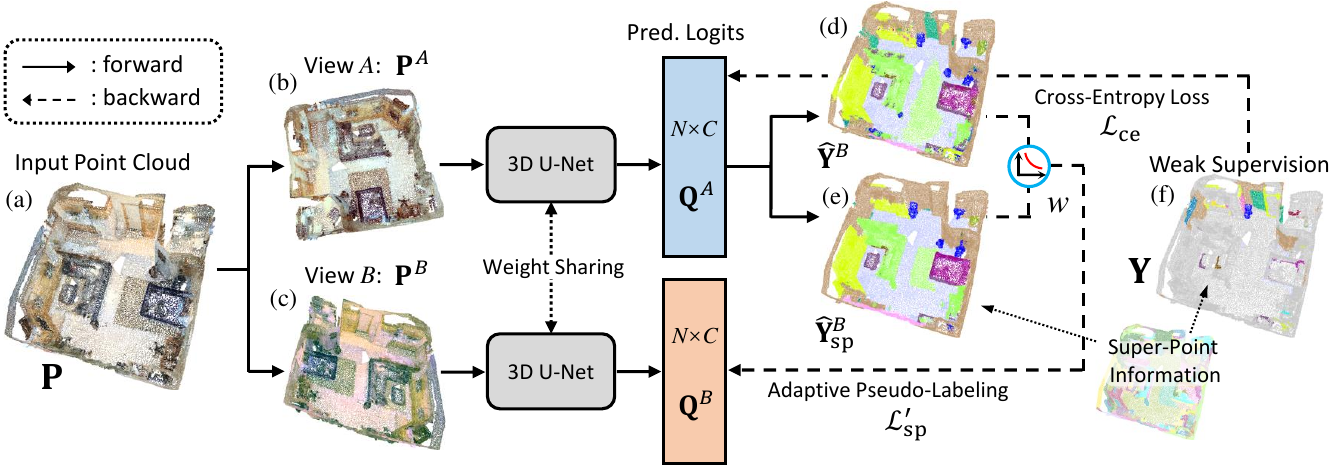}
\caption{The overview of PointMatch. (a) the input point cloud; (b) the view $A$ augmented from the input point cloud; (c) view $B$ generated via another augmentation; (d) the point-wise pseudo-label; (e) the super-point-wise pseudo-label; (f) the weak supervision (``1thing1click'' setting), points in gray are unlabeled ones and other colors indicate different semantic meanings. }
\label{fig:overview}
\vspace{-4mm}
\end{figure*}
\section{Related Work}
\paragraph{Fully Supervised 3D Semantic Segmentation} Semantic segmentation approaches for 3D point cloud can be mainly classified into two groups: point-based and voxel-based methods. Point-based Methods~\cite{qi2017pointnet, qi2017pointnet++, li2018pointcnn, wu2019pointconv, wang2019dynamic, thomas2019kpconv, lin2020fpconv} apply convolutional kernels to a local region of points for feature extraction and the neighbors of a point are computed from k-NN or spherical search. In the case of voxel-based methods~\cite{graham20183d, choy20194d, wang2019voxsegnet, huang2021supervoxel}, the points in the 3D space are first transformed into voxel representations so that standard CNN can be adopted to process the structured voxels. In either point-based or voxel-based methods, feature aggregation is performed in the Euclidean space, while there are some recent works~\cite{jiang2019hierarchical, lei2020spherical, schult2020dualconvmesh, hu2021vmnet} that consider geodesic information for better feature representation. More recently, the Transformer structure~\cite{zhao2021point} is also proposed for point clouds, as an alternative to the classic convolutional structure. However, most of the above methods are designed for the fully-supervised scheme, while annotation on point clouds is exhausting and costly, especially in the application of semantic segmentation, where the scene (indoor or outdoor) point cloud is usually of a large scale. In this work, we focus on weakly-supervised point cloud segmentation, where only very sparse points are annotated in each scene. 

\vspace{-2mm}
\paragraph{Weakly Supervised 3D Semantic Segmentation} Existing works explore the 3D semantic segmentation with various types of weak supervision, including 2D image~\cite{wang2019towards}, subcloud-level~\cite{wei2020multi}, segment-level~\cite{tao2020seggroup}, and point-level supervision~\cite{xu2020weakly, hu2021sqn, liu2021one, shi2021labelefficient}. The first three types can be grouped into indirect annotations~\cite{hu2021sqn}. The work of \cite{wang2019towards} utilizes the annotations on the projected 2D image of a point cloud, with only single view per sample. In \cite{wei2020multi}, a classifier is trained first with sub-cloud labels, from which point-level pseudo labels can be generated via class activation mapping techniques~\cite{zhou2016learning}. In another way, the work of \cite{tao2020seggroup} pre-generates segments/super-points to extend sparse click annotation into segment-level supervision, and groups unlabeled segments into the relevant nearby labeled ones for label sharing. For point-level weak supervision, the work of \cite{hu2021sqn} proposes to use only 10\% of labels by learning gradient approximation and utilizing low-level smoothness constraints. A harder setting with a much lower label ratio, 1\textperthousand, is further investigated in \cite{hu2021sqn}, where a Semantic Query Network (SQN) is proposed based on leveraging the semantic similarity between neighboring points. Another work OTOC~\cite{liu2021one} proposes a novel weakly supervised setting, One Thing One Click (``1thing1click''), \ie, with only one point annotated for each instance in the scene. They employ an extra branch of network to probe the relation between super-points and propagate labels among highly-related ones. Besides, authors of \cite{shi2021labelefficient} propose an active learning approach for annotating selected super-point with a limited budget to maximize model performance.
Another line of works is contributed to self-supervised pre-training of 3D point clouds~\cite{sharma2020self, liu2020p4contrast, xie2020pointcontrast, hou2021exploring, zhang2021self}. The pre-training usually needs weak or even no labels and provides a better network initialization for the downstream tasks. 

Existing point-level weakly supervised 3D semantic segmentation methods act on label propagation by leveraging the relation between points/super-points. However, the proposed PointMatch takes a novel way based on consistency regularization to better probe information in the point cloud data itself and alleviates the reliance on the given labels. 

\vspace{-2mm}
\paragraph{Consistency Training}
Consistency training is a powerful algorithmic paradigm proposed for robust learning from label-scarce data. It is constructed on enforcing the prediction stability under different input transformations~\cite{wei2020theoretical}, \eg adversarial perturbations~\cite{miyato2018virtual} or data augmentations~\cite{sohn2020fixmatch, xie2020self}, in the manner of pseudo-labeling, \ie, using the prediction of one transformation as the  fitting target of the other. Thus it combines the advantages of both consistency regularization and pseudo-labeling (or self-training). This approach has been applied in many domains, such as semi-supervised learning (SSL)~\cite{berthelot2019mixmatch, berthelot2019remixmatch, sohn2020fixmatch, xie2020self}, unsupervised learning (USL)~\cite{hu2017learning, grill2020bootstrap}, unsupervised domain adaptation (UDA)~\cite{french2017self, shu2018dirt}, and semi-supervised domain adaptation (SSDA)~\cite{li2021cross, li2021semi}, all of which prove the effectiveness of consistency training in learning high-quality representations from label-scarce data. More recently, there are some works extending consistency training into other tasks, such as unsupervised domain adaptation for image segmentation~\cite{melas2021pixmatch} and semi-supervised 3D object detection~\cite{wang20213dioumatch}. 

To our knowledge, it is the first time that consistency training is applied in the weakly supervised semantic segmentation of 3D point clouds. Different from the previous works, consistency training is novelly used in a weakly-supervised scenario where limited point-wise supervision is provided in each training sample. In addition, our work properly leverages the super-point grouping information in point clouds to further improve the whole framework. 
\section{Methodology}
\subsection{Problem Definition}
We first formulate the weakly supervised 3D semantic segmentation problem, taking the indoor scene scenario as an example. Given the point cloud $\mathbf{P} \in \mathbb{R}^{N\times D}$ of a scene of $N$ points with $D$-dimension features, there are only partial points annotated for training. The points with labels are denoted as $\{(\mathbf{x}_i^l, y_i), i \in L$\}, and other unlabeled points are denoted as $\{\mathbf{x}_i^u, i \in U\}$, where $L$ and $U$ are two sets, satisfying $L \cap U = \varnothing$ and $L \cup U = \langle N\rangle$ ($\langle N\rangle$ is a short form of $\{1, 2, \cdots, N \}$, the same hereinafter). The target of $f$ is to predict the semantic category $y_i\in \langle C\rangle$ of each point $\mathbf{x}_i$, where $C$ is the number of possible categories. Taking the point cloud $\mathbf{P}$ as input, $f$ outputs the prediction probability $\mathbf{Q} \in [0, 1]^{N \times C}$ over all $C$ classes, for all $N$ points of $\mathbf{P}$. Note that the summation of values in each row of $\mathbf{Q}$ is equal to 1. Denote the weak semantic label of the whole scene as $\mathbf{y} \in {\langle C\rangle}^N$ and its one-hot extension as $\mathbf{Y} \in \{0, 1\}^{N\times C}$. 
To optimize $f$, a straightforward way is to compute the cross-entropy loss $\mathcal{L}_{ce}$ between $\mathbf{Q}$ and $\mathbf{Y}$, formulated as:
\begin{align}
    \mathcal{L}_{ce} = \frac{1}{|L|}\sum_{i \in L} \text{cross-entropy}(\mathbf{Q}_i, ~\mathbf{Y}_i),
\label{eq:ce}
\end{align}
where $|L|$ represents the set size of $L$ and the subscript $i$ indicates the row index, so $\mathbf{Q}_i$ and $\mathbf{Y}_i$ are two $C$-class distributions corresponding to the $i$-th point. 
At the inference stage, the semantic segmentation result of a scene can be generated from $f$'s prediction, by simply choosing the class with the highest score in each row of $\mathbf{Q}$. 


To probe more information the limited labels and point cloud data itself, we design a novel framework, PointMatch, with the pipeline illustrated in  Fig.~\ref{fig:overview}. It conducts a consistency training framework designed for weakly labeled point clouds, and an adaptive pseudo-labeling mechanism by incorporating the super-point information, described in the following Sec.~\ref{sec:consist_train} and Sec.~\ref{sec:adaptive_psdo}, respectively. 

\subsection{Consistency Training}
\label{sec:consist_train}
The proposed consistency training framework focuses on better exploitation of  data itself, by encouraging the model's point-wise predictive consistency between two views of an input scene, through employing the prediction in one view as the pseudo-label of the other. 
Such a consistency training approach has three advantages: (i) various augmentations enables the network robust to different kinds of perturbation on low-level input features; (ii) the consistency target facilitates the model's ability in extracting high-level semantic features from the point cloud data itself; (iii) the self-training process implicitly propagates sparse training signals to unlabeled points and provide dense pseudo-labels, which increases the learning stability. 

Formally, given a point cloud $\mathbf{P} \in \mathbb{R}^{N\times D}$, our PointMatch applies two different groups of data augmentations to create its two views $\mathbf{P}^A \in \mathbb{R}^{N\times D}$ and $\mathbf{P}^B \in \mathbb{R}^{N\times D}$, respectively. To avoid breaking the local structure of the point cloud too much, we perform scene-level augmentations like offsetting, scaling, rotation, flipping, jittering, \etc. The obtained two views $\mathbf{P}^A$ and $\mathbf{P}^B$ are then fed into the 3D U-Net $f_{\theta}$ for point-wise semantic prediction, where $\theta$ is the network parameters. The network $f_{\theta}$ outputs the per-point probability distribution of $\mathbf{P}^A$, denoted as $\mathbf{Q}^A \in [0, 1]^{N\times C}$, and similarly, $\mathbf{Q}^B \in [0, 1]^{N\times C}$ can be generated from $\mathbf{P}^B$, formulated as:
\begin{equation}
\begin{aligned}
    \mathbf{Q}^A &= f_{\theta}(\mathbf{P}^A),\\
    \mathbf{Q}^B &= f_{\theta}(\mathbf{P}^B).
\end{aligned}
\end{equation}
In the next step, we generate the pseudo-label of $\mathbf{Q}^B$ from $\mathbf{Q}^A$ to create the self-consistency loop. Specifically, the most-likely predictive category of each point (as well as its confidence score) is chosen to form the pseudo-label, \ie, the indices of the highest value in each row of $\mathbf{Q}^A$. However, $\mathbf{Q}^A$ is usually noisy and even contains many uncertain predictions, so a direct use may provide negative guidance to $\mathbf{Q}^B$ and harm the whole learning scheme. Hence, we conduct a filtering operation to improve the pseudo-label quality, by ignoring those predictions with confidence lower than a threshold $\tau$. Denote the filtering mask as $\mathbf{m} \in [0,1]^{N}$, which is generated as follows:
\begin{equation}
\mathbf{m}_i = 
\left\{
\begin{array}{lr}
    1, ~~\max(\mathbf{Q}^A_i) \geq \tau,  \\
    0, ~~\text{otherwise},
\label{eq:mask}
\end{array}
\right.
\forall i \in \langle N\rangle,
\end{equation}
where $i$ is the row index of $\mathbf{Q}^A$ and $\tau$ is set as 0.95 in our implementation. 
Given $\mathbf{m}$ and the one-hot extension of $\mathbf{Q}^B$'s pseudo-label, represented as $\widehat{\mathbf{Y}}^{B} \in \{0, 1\}^{N\times C}$, the pseudo-labeling of $\mathbf{Q}^B$ can be conducted via a cross-entropy loss:
\begin{align}
    \mathcal{L}_{pl} = \frac{1}{N}\sum_{i \in \langle N\rangle} \mathbf{m}_i \cdot \text{cross-entropy}(\mathbf{Q}^B_i, ~\widehat{\mathbf{Y}}^{B}_i).
\label{eq:loss_pl}
\end{align}
Until this point, we are working on probing information only from point cloud data itself for a better data exploitation. Then the weak labels are integrated to provide discriminative semantic information, by using $\mathbf{Y}$ as the supervision of $\mathbf{Q}^A$ via computing a cross-entropy loss as in Eq.~\ref{eq:ce}. The parameters $\theta$ can then be optimized by minimizing the objective loss function $\mathcal{L}_{total}$ as follows: 
\begin{align}
    \min_{\theta} \mathcal{L}_{total} = \min_{\theta} \mathcal{L}_{ce} + \lambda \mathcal{L}_{pl},
\label{eq:loss_total}
\end{align}
where $\lambda$ is a scalar weight for balancing the two loss functions. As the learning process goes, the model exploits the knowledge learned from the limited annotations to train itself via forcing the predictive consistency, and meanwhile, implicitly propagates the sparse training signals to the whole scene via pseudo-labeling. 

\subsection{Adaptive Pseudo-Labeling}
\label{sec:adaptive_psdo}
Although the framework above facilitates the model's robust learning subtly, we observe that there are still considerable mispredictions in the obtained pseudo-labels, especially at the early learning stage. One reason is that the previous training scheme is mainly constructed on the predictive consistency between each pair of single points, and the inter-point relation information is learned insufficiently. Therefore, we further exploit the super-point prior to introduce local structure information of point cloud for generating pseudo-labels of higher quality. 

The super-points of a scene can be generated via an unsupervised low-level clustering by the position and color information of each point. We refer to \cite{landrieu2018large} for the manner of super-point generation, and it is recommended for more details. Formally, given a point cloud $\mathbf{P} \in \mathbb{R}^{N \times D}$, we obtain a set of super-points $\{\mathbf{S}^{(i)}\}, i \in \langle M \rangle$, where $M$ is the number of super-point and each $\mathbf{S}^{(i)} \in \mathbb{R}^{S^{(i)} \times D}$ includes $S^{(i)}$ $D$-dimension points. Each point in $\mathbf{P}$ belongs to one super-point only, so $\mathbf{S}^{(i)} \cap \mathbf{S}^{(j)} = \varnothing, \forall i \neq j$ and the summation of all $S^{(i)}$ is equal to $N$. The obtained super-point information is then used to improve the quality of point-wise pseudo-label $\widehat{\mathbf{Y}}^{B}$. Given point-wise predictions in each super-point group, a voting operation is carried out to get a ``mainstream'' category. The elected category is then propagated to all points in this group to obtain a super-point-wise pseudo-label $\widehat{\mathbf{Y}}^{B}_\text{sp}$. An illustrative example of  $\widehat{\mathbf{Y}}^{B}$ and $\widehat{\mathbf{Y}}^{B}_\text{sp}$ is shown in Fig.~\ref{fig:overview} (d) and (e), respectively. It can be observed that $\widehat{\mathbf{Y}}^{B}_\text{sp}$ tends to have higher purity. Similar to Sec.~\ref{sec:consist_train}, we preserve confident predictions to form high-quality super-point-wise pseudo-labels. Specifically, given $\mathbf{Q}^B$, the average probability distribution in each super-point is computed first, of which the category with the highest score is selected and propagated in the whole super-point. Then the filtering mask $\mathbf{m}^\text{sp}$ is generated by checking whether the confidence of each point is beyond a pre-defined threshold $\tau^\text{sp}$, similar to the computation in Eq.~\ref{eq:mask}. 

Although the voting operation enables $\widehat{\mathbf{Y}}^{B}_\text{sp}$ more stable and accurate, it suffers from the inherent noise arising from the super-point generation process. Thus, the point-wise pseudo-labels may have higher accuracy when the model is strong enough. Accordingly, we further design an adaptive combination mechanism to exploit the advantages of both. At the early stage, the learning of $f_\theta$ relies on $\widehat{\mathbf{Y}}^{B}_\text{sp}$ via a cross-entropy loss $\mathcal{L}_{pl}^\text{sp}$:
\begin{align}
    \mathcal{L}_{pl}^\text{sp} = \frac{1}{N}\sum_{i \in \langle N\rangle} \mathbf{m}^\text{sp}_i \cdot \text{cross-entropy}(\mathbf{Q}^B_i, ~\widehat{\mathbf{Y}}^{B}_{\text{sp}i}).
\end{align}
As the learning goes, an adaptive weight $w$ is adopted to gradually incorporate $\mathcal{L}_{pl}$ (Eq.~\ref{eq:loss_pl}) and abandon $\mathcal{L}_{pl}^\text{sp}$:
\begin{align}
    \mathcal{L}_{pl}' = w \cdot \mathcal{L}_{pl}^\text{sp} + (1-w) \cdot \mathcal{L}_{pl},
\end{align}
where $w$ is a scalar in the range of $[0,1]$ and drops from 1 to 0 gradually with an inverse decay. Formally, the adaptive weight $w$ at the $k$-th training epoch can be computed as:
\begin{align}
    w = \alpha \cdot k^{-1}, k \in \mathbb{N},
\label{eq:w}
\end{align}
where $\alpha > 0$ indicates the decay ratio.
In this way, at the late stage of training, $f_\theta$ can be completely supervised by the point-wise pseudo-label, so that the model can keep from the noise in super-point grouping. The new pseudo-labeling loss $\mathcal{L}_{pl}'$ is used to substitute the original $\mathcal{L}_{pl}$ in Eq.~\ref{eq:loss_total} for the final loss function.

\section{Experiments}
\begin{table}[t]
    \centering
    \caption{ MIoU (\%) on the ScanNet-v2 dataset (online test set). * means the performance of our baseline on the fully-supervised setting. The underline indicates the previous SOTA performance on each setting. The supervision types ``subcloud'' and ``segment'' mean using subcloud-level and segment-level annotation, respectively. ``20 points'' and ``1thing1click'' mean annotating 20 points per scene and annotating one point in each instance, respectively. }
    \resizebox{0.9\linewidth}{!}{
        \begin{tabular}{
        p{3.5cm}|
        >{\centering\arraybackslash}p{2cm}
        >{\centering\arraybackslash}p{2cm}
        }
        \toprule
        \textbf{~Method}
        & \textbf{Supervision}
        & \textbf{MIoU} \\  
        \midrule
        \cite{qi2017pointnet++}~PointNet++         & 100\% & 33.9 \\
        \cite{su2018splatnet}~SPLATNet             & 100\% & 39.3 \\
        \cite{tatarchenko2018tangent}~TangentConv  & 100\% & 43.8 \\
        \cite{li2018pointcnn}~PointCNN             & 100\% & 45.8 \\
        \cite{lin2020fpconv}~FPConv                & 100\% & 63.9 \\
        \cite{wu2019pointconv}~PointConv           & 100\% & 66.6 \\
        \cite{thomas2019kpconv}~KPConv             & 100\% & 68.4 \\
        \cite{choy20194d}~MinkowskiNet             & 100\% & 73.6 \\
        \cite{hu2021vmnet}~VMNet                   & 100\% & 74.6 \\
        \cite{han2020occuseg}~Occuseg              & 100\% & 76.4 \\
        \cite{nekrasov2021mix3d}~Mix3D             & 100\% & 78.1 \\
        \midrule
        \cite{graham20183d}~SparseConv             & 100\% & ~~72.5*\\

        \midrule[0.7pt]
        \cite{wei2020multi}~MPRM         & subcloud     & 41.1\\
        \cite{tao2020seggroup}~SegGroup  & segment      & 61.1\\
        \cite{hu2021sqn}~SQN             & 0.01\%       & \underline{35.9}\\
        \cite{hu2021sqn}~SQN             & 0.1\%        & \underline{51.6}\\
        \cite{liu2021one}~OTOC           & 20 points    & \underline{59.4}\\
        \cite{liu2021one}~OTOC           & 1thing1click & \underline{69.1}\\
        \midrule
        ~\textbf{PointMatch} & 0.01\%       & \textbf{57.1}\\
        ~\textbf{PointMatch} & 0.1\%        & \textbf{68.8}\\
        ~\textbf{PointMatch} & 20 points    & \textbf{62.4}\\
        ~\textbf{PointMatch} & 1thing1click & \textbf{69.5}\\
        \bottomrule
        \end{tabular}
    }
    \label{tab:scannet_test}
    \vspace{-3mm}
\end{table}
\subsection{Experiment Setup}
\paragraph{Datasets and Metric} We choose two popular point cloud datasets for the evaluation of our method, ScanNet-v2~\cite{dai2017scannet} and S3DIS~\cite{armeni20163d}. The ScanNet-v2 dataset~\cite{dai2017scannet} contains the 3D scans of 1,613 indoor scenes of 20 semantic categories (1,201 for training, 312 for validation, and 100 for online test). The whole dataset includes around 243 million points in total. The S3DIS dataset~\cite{armeni20163d} contains 271 room point clouds with 13 categories, scanned from 6 areas. Following the official train/validation split, Area 1,2,3,4,6 are used for training and Area 5 is used for evaluation. Besides, the S3DIS dataset has 273 million points, \ie, around 1 million points per scene on average, which is denser than scenes in the ScanNet dataset. The evaluation metric for 3D semantic segmentation we use is intersection-over-union, and we report the mean result (MIoU) over all categories for comparison with other methods. 

\vspace{-2mm}
\paragraph{Implementation Details}
We adopt SparseConv~\cite{graham20183d} as the 3D U-Net backbone in PointMatch. 
The output dimension of the SparseConv is set to 32, which is the same as in \cite{liu2021one}. Following \cite{jiang2020pointgroup} and \cite{liu2021one}, we randomly sample 250k points for too large scenes in the ScanNet-v2 dataset. We use two different combinations of various augmentations to create two views, randomly chosen from scaling, flipping, offsetting, rotation, affine transformation, position jittering, and color jittering, with a random augmentation extent. Hyper-parameters in our experiment $\tau$,  $\tau^\text{sp}$, $\epsilon$, $\lambda$, and $\alpha$ are set to 0.95, 0.95, 0.5, 1.0, and 1.0, respectively. The network is trained for 512 epochs using Adam optimizer~\cite{kingma2014adam} with a learning rate of 0.01 and a mini-batch size of 8 on the ScanNet-v2 dataset and 4 for the S3DIS dataset. Considering the total number of training epochs, we replace the epoch number $k$ in Eq.~\ref{eq:w} with $\lfloor k/64 \rfloor$ on the ``1thing1click'' setting and $\lfloor k/32 \rfloor$ on others, which is the round-off of $k$ divided by 32 or 64, in order to slow the decay rate. For the super-point generation, we follow \cite{liu2021one} to use the mesh segment results~\cite{dai2017scannet} on the ScanNet-v2 dataset and the super-point graph partition manner proposed by \cite{landrieu2018large} on the S3DIS dataset. Note that the super-points are used in training, and the inference stage does not rely on super-points. All experiments are conducted on an Intel Xeon Gold 6226R CPU and an NVIDIA RTX3090 GPU with 24GB memory.

\subsection{Experiment Results}
\begin{table}[t]
    \centering
    \caption{ MIoU (\%) on the ScanNet-v2 dataset validation set. * means the performance of our baseline on the fully-supervised setting. Note that SQN~\cite{hu2021sqn} reports only its performance of 0.1\% label setting on the ScanNet-v2 validation set. }
    \resizebox{0.9\linewidth}{!}{
        \begin{tabular}{
        p{3.5cm}|
        >{\centering\arraybackslash}p{2cm}
        >{\centering\arraybackslash}p{2cm}
        }
        \toprule
        \textbf{~Method}
        & \textbf{Supervision}
        & \textbf{MIoU} \\  
        \midrule
        \cite{graham20183d}~SparseConv   & 100\% & ~~72.2*\\

        \midrule[0.7pt]
        \cite{hu2021sqn}~SQN             & 0.1\%        & 53.5\\
        \cite{liu2021one}~OTOC           & 20 points    & 61.4\\
        \cite{liu2021one}~OTOC           & 1thing1click & 70.5\\
        \midrule
        ~\textbf{PointMatch} & 0.01\%       & \textbf{58.7}\\
        ~\textbf{PointMatch} & 0.1\%        & \textbf{69.3}\\
        ~\textbf{PointMatch} & 20 points    & \textbf{64.8}\\
        ~\textbf{PointMatch} & 1thing1click & \textbf{70.7}\\
        \bottomrule
        \end{tabular}
    }
    \label{tab:scannet_val}
    \vspace{-3mm}
\end{table}
\paragraph{Evaluation on ScanNet-v2}
On the ScanNet-v2~\cite{dai2017scannet} dataset, the evaluation of PointMatch is conducted on four weakly-supervised settings, \ie, 0.01\% of points annotated in each scene~\cite{hu2021sqn}, 0.1\% of points annotated in each scene~\cite{hu2021sqn}, 20 points annotated per scene~\cite{hou2021exploring} (20 points), and 1 point annotated for each instance in the scene~\cite{liu2021one} (1thing1click). The annotated points in the first two settings (0.01\% and 0.1\%) are randomly chosen following ~\cite{hu2021sqn}. The ``20 points'' setting is implemented following the officially ScanNet-v2 ``3D Semantic label with Limited Annotations'' benchmark~\cite{hou2021exploring}. Annotated points in the ``1thing1click'' setting are randomly chosen from each instance following \cite{liu2021one}. The average point label in this setting is around 0.02\%~\cite{liu2021one}. The evaluation results on the ScanNet-v2 online test set are presented in Tab.~\ref{tab:scannet_test}. Existing weakly supervised 3D semantic segmentation methods are also included for comparison, and some fully supervised methods are also listed in the table. As shown in the table, the proposed PointMatch consistently surpasses all existing methods over all weakly-supervised settings. It outperforms the state-of-the-art (SOTA) result by 21.2\% on the 0.01\% setting, by 17.2\% on the 0.1\% setting, and by 3.0\% on the ``20 points'' setting. The performance on the ``1thing1click'' setting is further close to the fully-supervised baseline. Note that the work OTOC~\cite{liu2021one} takes 5 turns of iterative training to reach the above results, which is around 1536 epochs (3 times of ours). In addition, we also provide the performance of PointMatch on the ScanNet-v2 validation set in Tab.~\ref{tab:scannet_val}, on four weakly-supervised settings mentioned above, which also proves the superiority of PointMatch. Detailed results over 20 categories are shown in supplementary materials. 

\vspace{-2mm}
\paragraph{Evaluation on S3DIS} 
\begin{table}[t]
    \centering
    \caption{ MIoU (\%) on the S3DIS dataset (Area-5 for validation). * means the performance of our fully-supervised baseline. The underline indicates the previous SOTA performance on each setting. }
    \resizebox{0.9\linewidth}{!}{
        \begin{tabular}{
        p{3.5cm}|
        >{\centering\arraybackslash}p{2cm}
        >{\centering\arraybackslash}p{2cm}
        }
        \toprule
        \textbf{~Method}
        & \textbf{Supervision}
        & \textbf{MIoU} \\  
        \midrule
        \cite{qi2017pointnet}~PointNet             & 100\% & 41.1 \\
        \cite{tchapmi2017segcloud}~SegCloud        & 100\% & 48.9 \\
        \cite{tatarchenko2018tangent}~TangentConv  & 100\% & 52.8 \\
        \cite{li2018pointcnn}~PointCNN             & 100\% & 57.3 \\
        \cite{landrieu2018large}~SPGraph           & 100\% & 58.0 \\
        \cite{choy20194d}~MinkowskiNet             & 100\% & 65.4 \\
        \cite{thomas2019kpconv}~KPConv             & 100\% & 67.1 \\
        \cite{zhao2021point}~PointTransformer      & 100\% & 70.4 \\
        \midrule
        \cite{graham20183d}~SparseConv             & 100\% & ~~63.7*\\

        \midrule[0.7pt]
        \cite{laine2016temporal}~$\Pi$ Model & 0.2\%        & 44.3\\
        \cite{tarvainen2017mean}~MT          & 0.2\%        & 44.4\\
        \cite{xu2020weakly}~DGCNN+CRF        & 0.2\%        & 44.5\\
        \cite{laine2016temporal}~$\Pi$ Model & 10\%         & 46.3\\
        \cite{tarvainen2017mean}~MT          & 10\%         & 47.9\\
        \cite{xu2020weakly}~DGCNN+CRF        & 10\%         & 48.0\\
        \cite{liu2021one}~OTOC               & 1thing1click & \underline{50.1}\\
        \cite{hu2021sqn}~SQN                 & 0.01\%       & \underline{45.3}\\
        \cite{hu2021sqn}~SQN                 & 0.1\%        & \underline{61.4}\\
        \midrule
        ~\textbf{PointMatch} & 1thing1click & \textbf{55.3}\\
        ~\textbf{PointMatch} & 0.01\%       & \textbf{59.9}\\
        ~\textbf{PointMatch} & 0.1\%        & \textbf{63.4}\\
        \bottomrule
        \end{tabular}
    }
    \label{tab:s3dis}
    \vspace{-3mm}
\end{table}
\begin{figure*}[t]
\centering
\includegraphics[width=0.91\linewidth]{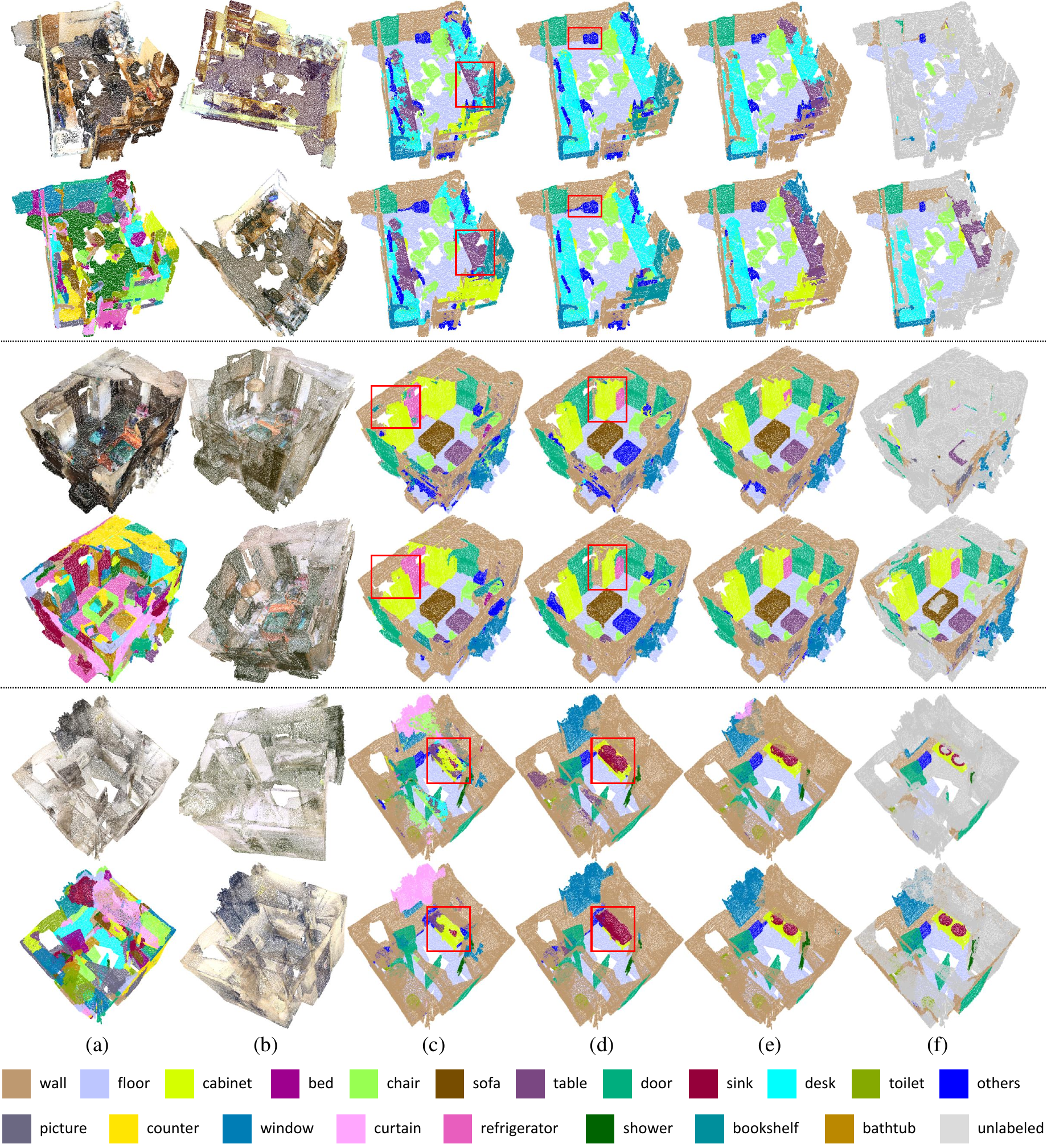}
\caption{Visualization of the qualitative results. We sample three scenes from the training set and their related results include, (a) upper: input point clouds, lower: the super-point grouping, in which colors do not indicate category information; (b): two views of the input point cloud; (c) upper: the point-wise pseudo-label at the early stage, lower: the super-point-level pseudo-label at the early stage; (d) upper: the point-wise pseudo-label at the late stage, lower: the super-point-level pseudo-label at the late stage; (e) upper: the weakly-supervised prediction, lower: the fully-supervised prediction; (f) upper: the weak supervision, lower: the full supervision (ground truth). }
\label{fig:vis_comp}
\vspace{-4mm}
\end{figure*}
We also evaluate the proposed method on the S3DIS~\cite{armeni20163d} dataset to further validate the effectiveness of the proposed method. Three weakly-supervised settings are included for evaluation, \ie, 0.01\%, 0.1\%, and  ``1thing1click'' (no official ``20 points'' setting provided for S3DIS). Note that the point cloud in the S3DIS dataset usually contains much more points than in the ScanNet-v2 dataset. By estimate, around 0.0036\% of points are annotated in the ``1thing1click'' setting. The results on these three settings are listed in Tab.~\ref{tab:s3dis}. The SOTA methods on both the fully-supervised and weakly-supervised settings are presented in the table for comparison. It is observed that the proposed PointMatch achieves the best performance over all three settings. It surpasses the SOTA result on the 0.01\% setting by a large margin of 14.6\%, by 5.2\% on the ``1thing1click'' setting, and by 2.0\% on the 0.1\% setting. Impressively, our result on the 0.1\% setting is very close to the fully-supervised baseline (63.4\% v.s. 63.7\%). The above results strongly prove the effectiveness and superiority of PointMatch, especially in the scenario of very sparse annotations (0.01\%). Detailed results on all 13 categories are listed in supplementary materials.  

\vspace{-3mm}
\paragraph{Qualitative Results}
\begin{table}[t]
    \centering
    \caption{ Ablative results of consistency training in PointMatch. MIoU (\%) on the ScanNet-v2 dataset validation set. }
    \resizebox{0.9\linewidth}{!}{
        \begin{tabular}{
        p{3.5cm}|
        >{\centering\arraybackslash}p{2cm}
        >{\centering\arraybackslash}p{2cm}
        }
        \toprule
        \textbf{~Method}
        & \textbf{Supervision}
        & \textbf{MIoU} \\  
        \midrule
        ~Fully-Sup. Version     & 100\%        & 72.2\\

        \midrule[0.7pt]
        ~PointMatch             & 0.01\%       & \textbf{58.7}\\
        ~w/o Consist. Training  & 0.01\%       & 51.3\\
        \midrule
        ~PointMatch             & 0.1\%        & \textbf{69.3}\\
        ~w/o Consist. Training  & 0.1\%        & 67.3\\
        \midrule
        ~PointMatch             & 20 points    & \textbf{64.8}\\
        ~w/o Consist. Training  & 20 points    & 55.0\\
        \midrule
        ~PointMatch             & 1thing1click & \textbf{70.7}\\
        ~w/o Consist. Training  & 1thing1click & 62.2\\
        \bottomrule
        \end{tabular}
    }
    \label{tab:abl_consist}
    \vspace{-3mm}
\end{table}
Except for the quantitative results, we also exhibit some qualitative segmentation results of PointMatch. As shown in Fig.~\ref{fig:vis_comp}, we visualize each sample in two rows and six columns, namely the input point cloud (upper) and its super-point grouping (lower) in column (a), its globally-augmented (upper) and locally-augmented (lower) views in column (b), its point-wise (upper) and super-point-wise (lower) pseudo-label at the early and late stage of training in column (c) and (d), respectively, the prediction of PointMatch under the weak (upper) and full (lower) supervision in column (e), and the corresponding weak label (upper) and ground truth (lower) in column (f). Note that all results we visualize are generated under the ``1thing1click'' weak supervision. It is observed that the predictions of PointMatch under weak supervision are close to the ground truths and the fully-supervised predictions. More impressively, the super-point-wise pseudo-labels are superior to the point-wise ones at the early stage, while get inferior at the late stage of training (see red boxes in Fig.~\ref{fig:vis_comp}), which confirms our claim. More visualization results are presented in supplementary materials for the space limit.   

\subsection{Ablation Study}
The proposed PointMatch mainly includes two components, the consistency training paradigm and the adaptive pseudo-labeling mechanism. Corresponding ablative experiments are conducted for the analysis of them. 

\vspace{-2mm}
\paragraph{Consistency Training}
To validate the effectiveness of the consistency training, we remove one branch in our framework as well as the pseudo-labeling mechanism, so the resultant version is a SparseConv simply trained on the weak supervision (extended by super-point information as the original) with a cross-entropy loss. We implement ablative experiments on four weakly-supervised settings on the ScanNet-v2 validation set. As shown in Tab.~\ref{tab:abl_consist}, removing the consistency training results in noticeable performance drops consistently over all weakly-supervised settings, especially on the schemes with extremely little supervision, which strongly proves its great effectiveness.

\vspace{-2mm}
\paragraph{Adaptive Pseudo-labeling}
The adaptive pseudo-labeling mechanism plays the role of pseudo-label correction at the early stage of training, and it is implemented with an inverse decay. To confirm the effectiveness of our design, we implement four versions on two weakly-supervised settings (``1thing1click'' and ``0.01\%'') for comparison: (i) using point-wise pseudo-label only ($w = \text{0}$); (ii) using super-point-wise pseudo-label only ($w = \text{1}$); (iii) using both two pseudo-labels but in a constant manner, by setting $w$ to 0.5 ($w = \text{0.5}$); (iv) using the adaptive mechanism with a larger decay ratio, by using $\lfloor k/32 \rfloor$ (``1thing1click'' settings) and $\lfloor k/32 \rfloor$ (0.01\% settings) in Eq.~\ref{eq:w} ($k \leftarrow \lfloor k/16(32) \rfloor$). Results are listed in Tab.~\ref{tab:abl_adaptive}. Using either type of pseudo-label only is inferior to the adaptive combination, because both point-wise and super-point-wise pseudo-label have their own strengths. Using a constant weight also leads to a performance drop, which proves that giving temporally different reliance on the two pseudo-labels can better exploit their advantages. Besides, a faster decay of the weight $w$ also results in a slightly worse result, which is usually close to the result of using point-wise pseudo-label only ($w = \text{0}$). One reason is that the network is unable to learn adequate information from super-points when $w$ drops too fast. 

\begin{table}[t]
    \centering
    \caption{ Ablative results of adaptive pseudo-labeling in PointMatch. MIoU (\%) on the S3DIS dataset Area-5. }
    \resizebox{0.9\linewidth}{!}{
        \begin{tabular}{
        p{3.5cm}|
        >{\centering\arraybackslash}p{2cm}
        >{\centering\arraybackslash}p{2cm}
        }
        \toprule
        \textbf{~Method}
        & \textbf{Supervision}
        & \textbf{MIoU} \\  
        \midrule
        ~Fully-Sup. Version     & 100\%        & 63.7\\

        \midrule[0.7pt]
        ~PointMatch         & 0.01\%       & \textbf{59.9}\\
        ~$w = \text{0}$     & 0.01\%       & 58.4\\
        ~$w = \text{1}$     & 0.01\%       & 56.1\\
        ~$w = \text{0.5}$   & 0.01\%       & 54.6\\
        ~$k \leftarrow \lfloor k/16 \rfloor$ & 0.01\%       & 58.7\\
        \midrule
        ~PointMatch         & 1thing1click & \textbf{55.3}\\
        ~$w = \text{0}$     & 1thing1click & 52.6\\
        ~$w = \text{1}$     & 1thing1click & 50.2\\
        ~$w = \text{0.5}$   & 1thing1click & 48.4\\
        ~$k \leftarrow \lfloor k/32 \rfloor$ & 1thing1click & 53.3\\
        \bottomrule
        \end{tabular}
    }
    \label{tab:abl_adaptive}
    \vspace{-3mm}
\end{table}

\section{Conclusion and Discussion}
We propose a novel approach, PointMatch, which introduces a consistency training framework into weakly supervised semantic segmentation of point clouds. It works by enforcing the predictive consistency between two views of a point cloud via pseudo-labeling, and enables the network to perform robust representation learning from weak label and data itself. The pseudo-label quality is further promoted by integrating  super-point information in an adaptive manner. Impressively, PointMatch achieves SOTA performance over various weakly-supervised semantic segmentation settings on both ScanNet-v2 and S3DIS datasets, and shows strong robustness given even extremely little labels, \eg 20 points per-scene and 0.01\% of points annotated. 


{\small
\bibliographystyle{ieee_fullname}
\bibliography{main}
}

\end{document}